\documentclass{article}

\usepackage{arxiv}

\usepackage[utf8]{inputenc} 
\usepackage[T1]{fontenc}    
\usepackage{hyperref}       
\usepackage{url}            
\usepackage{booktabs}       
\usepackage{amsfonts}       
\usepackage{nicefrac}       
\usepackage{microtype}      
\usepackage{graphicx}
\usepackage{epsfig,color,amsmath,latexsym,amssymb}

\usepackage[linesnumbered,ruled,vlined]{algorithm2e}

\usepackage{multirow}
\usepackage{pgffor}
\usepackage{subcaption}

\def\bx{\mathbf{x}}
\def\RR{\mathrm{\hbox{I\kern-.2em\hbox{R}}}}
\def\eps{\varepsilon}

\newcommand{\ba}{\begin{array}}
\newcommand{\ea}{\end{array}}
\newcommand{\beq}{\begin{equation}}
\newcommand{\bal}{\begin{aligned}}
\newcommand{\ec}{\end{center}}
\newcommand{\eeq}{\end{equation}}
\newcommand{\eal}{\end{aligned}}

\author{
    Parisa ~Ghane \\
    Department of Electrical Engineering\\
    Texas A\&M University\\
  College Station, TX 77843 USA \\
  \texttt{pghane@tamu.edu}\\
  \And
  Gahangir Hossain \\
  Department of Computer Information Systems\\
  West Texas A\&M University\\
  Canyon, Texas 79016 \\
  \texttt{ghossain@wtamu.edu} \\
}

\title{Learning Patterns in Imaginary Vowels for an Intelligent Brain Computer Interface (BCI) Design}

\date{}

\begin{document}
\maketitle

\begin{abstract}
Technology advancements made it easy to measure non-invasive and high-quality electroencephalograph (EEG) signals from human's brain. Hence, development of robust and high-performance AI algorithms becomes crucial to properly process the EEG signals and recognize the patterns, which lead to an appropriate control signal. Despite the advancements in processing the motor imagery EEG signals, the healthcare applications, such as emotion detection, are still in the early stages of AI design. In this paper, we propose a modular framework for the recognition of vowels as the AI part of a brain computer interface system. We carefully designed the modules to discriminate the English vowels given the raw EEG signals, and meanwhile avoid the typical issued with the data-poor environments like most of the healthcare applications. The proposed framework consists of appropriate signal segmentation, filtering, extraction of spectral features, reducing the dimensions by means of principle component analysis, and finally a multi-class classification by decision-tree-based support vector machine (DT-SVM). The performance of our framework was evaluated by a combination of test-set and resubstitution (also known as apparent) error rates. We provide the algorithms of the proposed framework to make it easy for future researchers and developers who want to follow the same workflow.   
\end{abstract}


\section {Introduction}
Communicating through external links, like wires and computers, has turned to be an interesting research area since after successful investigation of brain's electrical activities, known as electroencephalography (EEG), by German neurologist Hans Berger \cite{haas2003hans}. Today, this communication system is called Brain Computer Interface (BCI) and is involved with the measurement and assessment of EEG around an individual’s brain. In particular, non-invasive EEG recording has recently become more popular due to the advancements in wireless technologies and electrical sensors as well as low cost and ease of use. To this end, it becomes crucial to develop AI algorithms and machine learning models that process the EEG measurements  and discover the potential patterns that may eventually lead to an appropriate control signal. 

Throughout the past couple of decades, extensive researches have been conducted on different aspects and applications of EEG-based BCIs. In 2019, \cite{wan2019review} studied the history of some successful EEG experiments and discussed discoveries about electrode placements as well as some EEG-based BCI paradigms. In the same year, \cite{abiri2019comprehensive}, provided a comprehensive analysis of the advantages and disadvantages of the previously proposed  EEG-based BCIs, with regards to the population under the study, utilization of algorithms, classification methods, and generation of control signals. They also provided suggestions to address the potential issues that may happen with each of the EEG-based BCI systems in the  aforementioned categories. Recently, \cite{gu2020eeg} highlighted the most popular EEG recording devices and signal/image processing methods, including filters, feature extractors, and classification rules, particularly those that originate from multi-layer perceptrons and neural networks classifiers. Although deep neural networks and transfer learning techniques have shown success in many other classification tasks, their performance is still unpredictable in some data-poor BCI applications, as opposed to the traditional methods \cite{lotte2018review}. 

Despite the EEG-based BCI advancements in some applications like controls, robotics, games and entertainment, the hardware and software seem to be not yet fully developed for healthcare, where the patients are the direct users of the BCI systems and EEG recording sessions are more challenging to set up. As a natural consequence, healthcare BCI applications face the lack of available EEG data while highly variable between participants. Accordingly, it remains an active research to design effective and efficient pattern recognition algorithms that suits the healthcare applications such as assessments of mental disorders, consciousness, emotions, communication impairments, and speech recognition \cite{huang2019eeg, pan2016eeg, ikeda2012effect, lee2019towards, krishna2020understanding}. 

Along the line of communication and emotion detection \cite{lee2019towards, krishna2020understanding, zhang2020eeg}, our research mainly contributes towards the potential BCI designs for vocal prosthesis by focusing on recognition of English vowels /a/, /e/, /i/, /o/, /u/. This will be particularly helpful for severely disabled people who do not have spontaneous reflexes in response to unexpected stimulation. The impulsive reactions, which are typically expressed in sounds, are the natural human response to environmental  stimulation, and without this ability paralyzed people may suffer from excess pain. In addition, a tool that reports the impulsive emotions, like surprise, grief, joy, and pain can be beneficial for both healthcare providers and patients during the treatment or a peak moment. Readers are referred to \cite{waaramaa2008monopitched} for more on the frequency and dominant vowels in expressing different emotions are. 

The EEG-based BCIs for vowel recognition have been under studies with a variety of perspectives. Some examples are as follows:\cite{yoshimura2011usability} developed a Bayesian network to detect phonemes with vowels /a/ and /u/. \cite{dasalla2009single} studied the imagery speech of vowels /a/ and /u/. \cite{wang2013extending} demonstrated that the ability of BCI for motor imaging can be enhanced by integrating motor imaging and speech imaging. \cite{zhang2020eeg} did a study on the Mandarin tone in two visual-only and audio-visual conditions, and showed that the latter condition resulted in a more accurate classification of imaginary Mandarin tones. 

In this paper, we present a modular framework for the processing of the EEG signals that had been previously recorded in the national university of Colombia. The whole model consists of pre-processing, feature extraction, dimensionality reduction, and classification modules. 
Our framework deals with the small sample size situation by carefully extracting only a few features and keeping the degrees of freedoms (i.e. unknown parameters) as low as possible. The former is recommended to avoid the curse of dimensionality \cite{trunk1979problem, hughes1968mean}, whereas the latter is required to avoid over (or under) fitting. In section 2 we briefly talk about the source of the data of this work and the experimental conditions for the EEG recording sessions. Section 3 explains the methods of this project. Results and discussions are presented in section 4, and finally section 5 gives the concluding remarks.  
\section{data acquisition}
The bio-electric traits of brain neurotransmitters are extensively described in \cite{malmivuo1995bioelectromagnetism, Sarmiento2019recognition}. For this work, we used the raw EEG data that had been recorded in controlled noise Clinical Electrophysiology Laboratory at National University of Colombia. The data collection experiments was described in \cite {sarmiento2014brain}, but we give a brief illustration here. The EEG signals were captured with a sampling frequency of 500 Hz, through 21 electrodes placed on the Wernicke and Broca areas, a.k.a language perception area \cite{pei2011spatiotemporal}. The EEG signals were recorded from 20 subjects (17 men and 3 women) while intending/imagining a vowel without moving the oral cavity or jaws. Each recording session consisted of 10 trials, each of which include 3 seconds of relaxation and 3 sends of imagination, for a total of 6 seconds per trial and one minute per session. A period of half-second was assumed to be the transition state between relaxation and imagination, and was later removed from the signals during the preprocessing step. For each trial, the data from all 21 channels together was imported to MatLab.
\section{Methods}
 In contrast to many other pattern recognition studies, we apply a few simple, though effective, preprocessing steps, and extract as few as possible  features. In pattern recognition problems, where the system is trained without any prior information, the simplicity of preprocessing and lack of features may cause week classification performance. However, for this EEG study, we were able to incorporate the prior biological and experimental knowledge, and consequently, design our preprocessing and feature extraction steps accordingly. Algorithms \ref{alg:preProc-featExt-dimReduc}, \ref{alg:dt}, \ref{alg:dt-svm} summarize the workflow of the proposed AI module of the EEG-based BCI system. 

\subsection {Preprocessing}
The preprocessing steps were similar to those in \cite{sarmiento2014brain, ghane2015silent} and were as follows.
The signal from each channel was separated from the rest, and then was broken into trials. The relaxation and transition states were removed from each trial and the remaining piece of signal was stored as a data sample. The stored signals from all sessions created a data set $X$ with total of 200 samples (for 10 trials per subject, and 20 subjects). A response variable $y \in \{1, 2, 3, 4, 5 \}$ was generated to refer to vowels /a/, /e/, /i/, /o/, /u/, respectively. Then the potential noise and artifacts were filtered out of the signal by applying a band-pass filter (BPF) with cut-off frequencies 2 and 50 hertz. We set lower band frequency to 2 to remove cardiac and ocular artifacts as well as muscle and electrode movements that may have been recorded during the sessions. The upper band frequency of 50 hertz filters the noise originating from the electrical power supply of devices in the recording room, which are in the range of 50-60 Hertz. Finally, the signals of all participants for each trial were normalize around mean and standard deviation. 

\subsection{Feature Extraction}
In this part, we carried a two-step feature extraction: the first step to extract the dominant spectral information, and the second one to reduce the dimensionality. 
Since the EEG recordings were done in short trials, in controlled situation and subjects were focused on the task, we assumed that the EEG signals in each trial is stationary (i.e. does not depend on time), and therefore expect the transformations based on Fourier transform carry important information to the frequency domain. 

The Fourier transform of the signal's auto-correlation function, known as power spectral Density (PSD) is a measure of the strength of a signal over a range of frequencies\cite{stoica2005spectral}, and is defined as \ref{eq:psd}. 
\beq
S_{xx}(\omega) = \int_{-\infty}^{+\infty} R_{xx}(\tau) e^{-i\omega \tau} d\tau
\label{eq:psd}
\eeq
 where $x$ represents the signal, $f$ is the frequency, $\omega= 2\pi f$, and $R_{xx}(\tau)$ is the auto-correlation function. Since the form of signal $x$ is not known, we empirically estimated the PSD with $\hat{S}_{xx}(\omega)$ using the periodogram with rectangular windows.

 Let's assume signal $x$ is discretized into $x_n$ by rectangular windows with size $\Delta t$. Then $x_n= x (n \Delta t)$ for $n= 0, 1, \dots, N-1$, and \ref{eq:psd} can be estimated by:
\beq
\hat{S}_{xx}(\omega) = \frac{\Delta t}{T} \left| \sum_{n=0}^{N-1} x_n e^{-i\omega n \Delta t}  \right|^2
\label{eq:perio-psd}
\eeq \\

\subsection{Dimensionality Reduction}
Once the power spectral densities were extracted, we applied principle component analysis (PCA) to map the data into $d$ orthogonal dimension, a.k.a. principle components (PCs). 
PCA is a dimensionality reduction method, originally proposed by \cite{pearson1901liii}, with proven applications across many disciplines \cite{shaghaghian2021studying, ghane2015robust,shlens2014tutorial}. 
The PCs are linear combination of original dimensions, and are sorted from the highest to the lowest ability in retaining the information of the original data. We took only the first two PCs ( total explained variance > \%75) for each channel, and then vectorized the features. Therefore, the feature space was reduced to 40 dimensions. Note that among 21 EEG electrodes, one was the reference electrode, and 40 dimensions was the result of the first two PCs of  PSD of the other 20 channels.

\subsection{Classification}
Support Vector Machines (SVM) has been widely used for BCI applications, in particular for binary class problems. In 2007, \cite{lotte2007review} highlighted that in terms of performances, SVM often outperformed other classifiers. Still after 10 years, SVM is one of the state of the art classification algorithms due to its flexibility and optimality, particularly in small sample sized data. SVM is a binary-class classification algorithm. We used a dendrogram-based algorithm \cite{takahashi2002decision} to extend the SVM to our multi-class BCI, and a created decision tree based multi-class SVM (DT-SVM). This was advantageous to one-against-one and one-against-all approaches \cite{vapnik2013nature, kijsirikul2002multiclass}, at least in terms of computational time. To distinguish N classes, the one-against-one approach requires to train a binary classifier on every combination of 2 classes, which makes total of  $N(N -1) / 2$ classifiers, whereas in a one-against-all approach, each class is separated from the rest by a single binary classifier and therefore total of $N$ classifiers are required.

Takahashi and Abe \cite{takahashi2002decision} proposed four alternatives to create the dendrogram for a multi-class problem: at each step, separating one class (or some classes) from the rest using either Euclidean distance or Mahalanobis distance. We used type 2, which is, at each step, separating some classes from the rest using the Euclidean distance, We put this method in algorithm algorithm \ref{alg:dt} for easier future implementation. The number of required classifiers depends on the separability of the classes,and is at most $N-1$. This is similar to the agglomerative hierarchical clustering (the bottom-up strategy), which groups the similar observations as it moves from the bottom to top of the dendrogram. 

\subsection{Performance Evaluation}
The performance of the system can be assessed by computing the classification error, which is the probability of misclassification. When the distribution of the data is not known, the error should be estimated by empirical measures. In the rest of this section, we propose using a convex combination of test-set error and resubstiution-like error to evaluated the performance of the system, experimentally.

In general, a test-set error on a large enough test set $S^{(t)}= (X^{(t)}, Y^{(t)})$ is regarded as the true error rate, and is a measure of generalization ability of the system. In binary classification problems, the test-set error is the sum of false negative and false positive. Ideally, the train and test data sets ($S_n$ and  $S^{(t)}$, respectively) should be from the same distributions, but disjoint (i.e. $S_n \cap S^{(t)} = \emptyset$) and independent (i.e. $P(S_n \cap S^{(t)}) = P(S_n) P(S^{(t)})$). 

Due to the lack of data samples, and non-linearity of the SVM classifiers, it was not feasible to set aside a sufficiently large test data set. Indeed, as much data as possible should be used for training of he system. In such data-poor environment we need to train and test the system on the same data set. Resubstitution and re-sampling methods are two well-known families of error estimators that test the system based on the training data. Resubstitution is computationally fast and not randomized, but is optimistically biased. On the other hand, re-sampling methods like cross validation are almost unbiased, but are randomized, and therefore not robust for small-sample-size situations (i.e. high variance) \cite{govindarajan2010evaluation, andersson1999measure}.

To deal with this situation, we set aside a small test set and estimated the error as a convex combination of an small test set error and a resubstitution-like error, when the resubstitution-like error. This is equivalent to creating a new test-like set $S^*=((X^{(t)}, Y^{(t)}) \cup (X^*, Y^*))$, where $(X^*, Y^*) \subset S_n$. The test-like error can be computed as:
\beq
\hat{\eps}_{(S^*)} = \frac{1}{m+n'} \sum_{(\bx, y) \in S^*} \left( I_{\psi(\bx)=1} I_{y=0} + I_{\psi(\bx)=0} I_{y=1} \right)
\label{eq:err}
\eeq
where $I_A$ is an indicator function defined as $I_A=0$ if condition $A$ is false and $I_A=1$ otherwise. $m$ is the size of the original $S^{(t)}$, and $n'<<n$ is the size of sample pairs $(X^*, Y^*)$. We suggest $n' \approx m$ to keep the balance between contribution of the test-like and the resubstitution-like error rates.


\begin{algorithm}
\caption {Pre-processing, feature extraction (PSD), and dimensionality reduction (PCA)} 
 \label{alg:preProc-featExt-dimReduc}
\KwIn{Raw EEG signals}
$n_s = $ total number of recording sessions\\
$n_t = $ total number of trials per session\\
$n_{ch} = $ total number channels\\
\ForEach {$s \in \{1, 2, \dots, n_s\}$,} {
	\ForEach {$j \in \{1, 2, \dots, n_t \}$, } {
	$trial_j= j^{th}$ trial in session $s$\\
	\ForEach{$i \in \{1, 2, \dots, n_{ch} \}$,} {
		$E=$ EEG portion of $trial_j$ (i.e. removing relaxation and transition states)\\
		$E^{(f)} = BPF(s)$ \\
		$\hat{S}=$ periodogram-PSD($E^{(f)}$) \ \ {\it \small (eq. \ref{eq:perio-psd})}\\
		$PCs = PCA(\hat{S})$\\
		$\bx[i, i+1]= [PC1 , PC2]$\\
		}
		$X_{ch}[j, :]= \bx$		{\it \small \ \ (set of signals per channel)}\\
	}
	$X[s, :]= X_{ch}$
}
 \KwOut{$X \in R^{(n_s \times n_t) \times 2n_{ch}}$.}
\end{algorithm}

\begin{algorithm}
\caption {Finding the dendrogram based on the Euclidean distance} 
 \label{alg:dt}
\KwIn{$X \subset R^{n\times d}$, $y \in \{0, \dots, K-1\}$}
\ForEach {class $i$, $i \in \{1, 2, \dots, K-1\}$,} {
$n_i = I_{y=i}$\\
$X_i = (X|y=i) $ \\
$\mathbf{c}_i= \frac{1}{n_i} \sum_{\bx \in X_i} \bx$
 }
 $C= \{ \{ 0 \}, \{ 1 \}, \dots, \{K-1\} \}$\\
 $L= \emptyset$  {\it \small (set of group of classes in the dendrogram.)} \\
 \ForEach{class $j \in C$}{
 $i= \underset{i \in C, i \neq j} {\mathrm{argmin}}  ||\mathbf{c}_i - \mathbf{c}_j||_2 $ \\
 $L= L \cup \{j, i \}$\\
 $C= C\backslash \{ j \} \backslash \{ i \} \cup \{ j, i \}$\\
 $X_i= [X_i , X_j]$\\ 
 $\mathbf{c}_i = \frac{1}{n_i} \sum_{\bx \in X_i} \bx$\\
 }
 \KwOut{$L$.}
\end{algorithm}

\begin{algorithm}
\caption {DT-SVM} 
 \label{alg:dt-svm}
\KwIn{$X \subset R^{n\times d}$, $y \in \{0, \dots, K-1\}$}
$L=$ output from algorithm \ref{alg:dt}\\
$\psi^{(DT)}= \emptyset$ \ \ \ \ \  { \small \it (set of trained classifiers)}\\
\ForEach {$i \in \{ 0, \dots, length(L) \} $, } {
	$l_i = L[i]$\\
	train $\psi_i$ on groups $l_i[0]$ and $l_i[1]$\\
	$\psi^{(DT)}= \psi^{(DT)} \cup \psi_i$
}
 \KwOut{$\psi^{(DT)}$}
\end{algorithm}

\section {Results and Discussion}
In this section, we describe step by step implementation of the algorithms and methods of this paper. 

\begin{figure}
    \centering
    \includegraphics[width= 0.9 \textwidth] {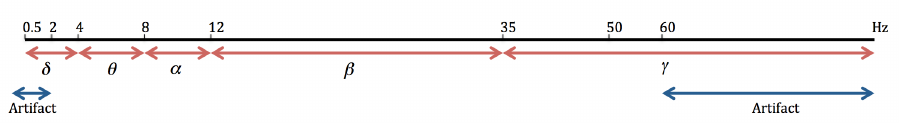}
    \caption{The range of frequency that was present in the EEG signals, which can be arrtibuted to bran waves ($\delta, \theta, \alpha, \beta, \gamma$) or the non-brain sources (artifacts with low frequencies like muscle and cardiac or high frequencies like electrical noise). Since the experiments were done with subjects being awake and fully relaxed, we expect the vowel-related signals be in $\alpha$ and $\beta$ ranges.}
    \label{fig:artifact}
\end{figure}

\begin{figure}
    \centering
    \includegraphics[width= 0.9 \textwidth] {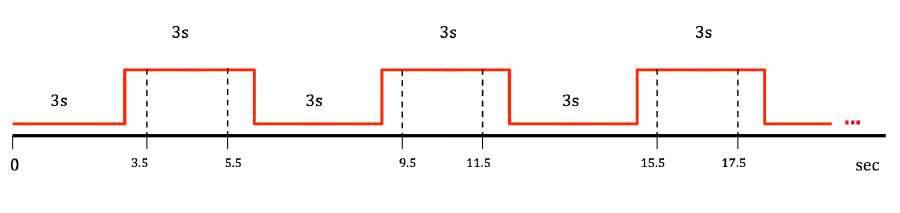}
    \caption{The relaxation (rest) and imagination (active) phases in one session: subjects were asked to repeatedly for 10 times do relax for 3 seconds and then start imagining a pre-specified vowel (active). The first and last half second of each active trial was removed, and the rest was taken as the signal of interest for later processings.}
    \label{fig:segment}
\end{figure}

The data includes total of $n_s=100$ recording sessions, for 5 vowels and 20 subjects. Each session's data was imported into MatLab in the form of a three dimensional matrix, with dimensions being time, electrode channel, and measured voltage. For each matrix, we separated the sessions into $n_t= 10$ trials, removed the non-brain artifacts including low-frequency components and high frequency noise (see figure \ref{fig:artifact}), as well as the relaxation and transition states, and retained only the 1000 samples that were captured between the remaining two seconds of active vowel imagination (see figure \ref{fig:segment}). (Note that the sampling frequency of the recording was 500Hz). Subsequently, we followed steps 7 through 12 of algorithm \ref{alg:preProc-featExt-dimReduc}, using the BPF, and the periodogram-PSD default setting in MatLab. 

The output of the algorithm \ref{alg:preProc-featExt-dimReduc} was the feature matrix $X \in R^{(n_s \times n_t) \times d}$ containing the $d=40-$dimensional feature vectors from $n_s= 20$ sessions, with $n_t= 10$ trials per session. We removed the first and last trials (trials 1 and 10) from all subject, because a large difference with other trials was observed within some subject, potentially because of not-fully-relaxed starting and fatigued ending.
Then among the remaining 8 trials per session, we extracted the data from trial 4 and set it aside as the test data, which made a test data set with $m= 100$ samples. The rest (700 samples) was considered as the train data $S_n$. Then we shuffled the training data once, picked the first $n'= 100$ first samples, and crated the test-like data set $S^*$ with size $m+n'=200$. 

\begin{figure}
    \centering
    \includegraphics[width= 0.9 \textwidth] {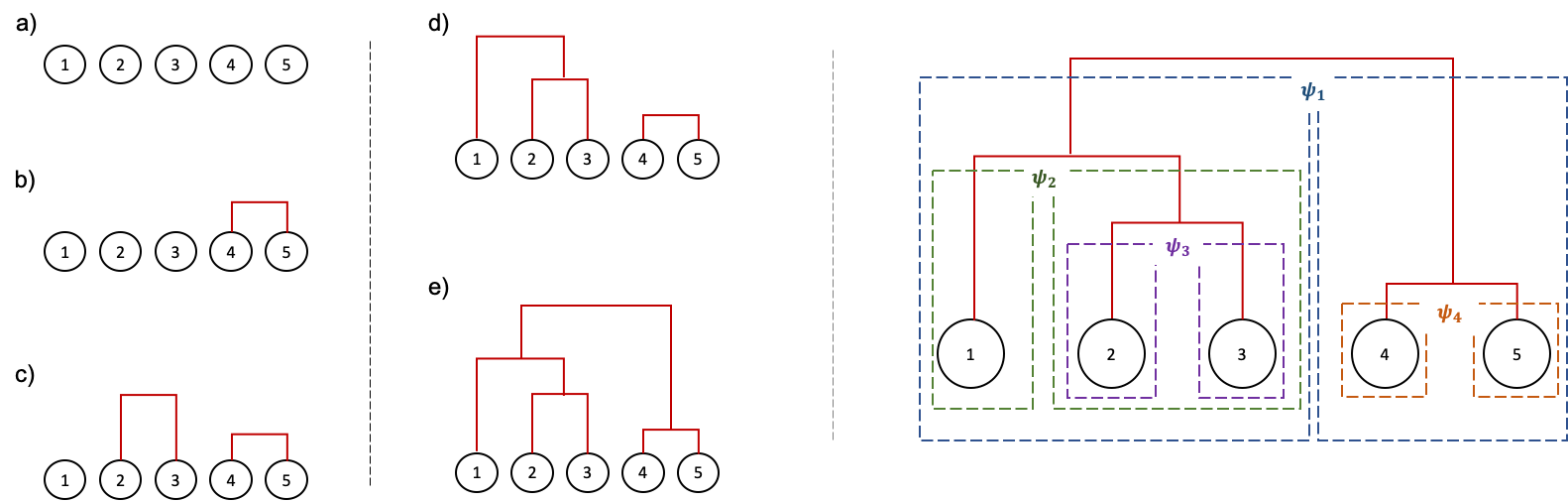}
    \caption{{\bf Left:} steps of dendrogram construction that was implemented as algorithm \ref{alg:dt}: a) original classes of the data, where labels /a/, /e/, /i/, /o/, and /u/ changed to integers 1, 2, 3, 4, and 5, respectively. b) Classes 4 and 5 (equivalently /o/ and /u/) were found to have the lowest Euclidean distance. c) next, comparing the Euclidean distance between classes 1, 2, 2, and the merged group of {4,5}, classes 2 and 3 were found to be closest. d) similarly, class 1 was merged to the joint group of {2 ,3}. e) the final (and only left) branch completes the tree. {\bf Right:} Each $\psi_i$ represents a binary SVM classifier at the $i^{th}$ level of the tree from top to bottom. As moving tho the bottom, $\psi_i$ becomes more specific about discriminating the pure classes.}
    \label{fig:DT-SVM}
\end{figure}

The training data $S_n$ was sent to algorithm \ref{alg:dt-svm}, where algorithm \ref{alg:dt} is called in the first line. Algorithm \ref{alg:dt} takes the feature-labeled data set, constructs  the dendrogram, and returns the structure of the dendromgram in the form of a set $L$. Figure \ref{fig:DT-SVM}, left, shows the steps of the dendrogram construction on our training data set. The corresponding set $L$ from algorithm \ref{alg:dt} was $L= \left\{ \{4, 5\} , \{ 2, 3 \}, \left\{ \{ 2, 3\}, \{ 1 \} \right\}, \left\{ \left\{ \{ 2, 3\}, \{ 1 \} \right\}, \{ 4, 5 \}	\right\}	 \right\}$, where the $i^{th}$ element of $L$ corresponds to the $i^{th}$ level of the dendrogram. For example, let's show the $i^{th}$ element of $L$ by $l_i$, then $l_1= \{ 4, 5 \}$ puts class 4 and 5 in one group in the first level, and $l_4= \left\{ \left\{ \{ 2, 3\}, \{ 1 \} \right\}, \{ 4, 5 \}	\right\}$ shows that the groups of $\{ 4, 5 \}$ and $ \left\{ \{ 2, 3\}, \{ 1 \} \right\}$ are merged in the $4^{th}$ and last level. Proceeding with algorithm \ref{alg:dt-svm}, a binary SVM classifier was trained to separate the the two branches at each level of the tree, as illustrated by figure \ref{fig:DT-SVM}, right. We used radial basis function (RBF) kernels for the SVM classifiers. The RBF kernel for sample $i$ and $j$ is defined as $k(\bx_i, \bx_j)$ in equation \ref{eq:rbf}, where $\sigma$ is a hyper-parameter that allows to optimize the classifier given the data
\begin{equation}
   k(x_i, x_j) = \exp \left(    \frac{||\bx_i, \bx_j||^2}{2\sigma^2}     \right)
   \label{eq:rbf}
\end{equation}

We evaluated the performance of the system by predicting the class label $y$ for the test-like data set $S^*$, and computed the test-like error rate by equation \ref{eq:err}. Table \ref{tab:conf_mat} is the confusion matrix that represents the result of the performance evaluation. The diagonal elements of the confusion matrix show the correctly classified rates, whereas the non-diagonal elements represent the misclassification rates. Let's use $M_{ij}$ to refer to the element on the $i^{th}$ row and $j^{th}$ column of the confusion matrix. The value of $M_{ij}$ represents the percentage that a sample from class $i$ was mistakenly placed by the classifier to class $j$. For example, $M_{12} = 0.05$ means that $5\%$ of the samples from class 1 (/a/) were predicted to be from class 2 (/e/). \\

Overall prediction error rate is 0.23 (equivalent to accuracy of 76.6\%. The highest and lowest accuracy was associated with vowel /a/ (90\%) and /i/ (65.5\%), respectively, as shown is table \ref{tab:acc}. Vowel /e/ was the most incorrectly predicted label for the data samples that came originally from classes /u/, /i/, and /a/, as can be seen in table \ref{tab:misclass}. The vowel /e/ itself was mostly incorrectly placed in class ($\psi=$) /a/, with 12.5\% misclassification. Interestingly, none of the vowels were incorrectly classified as ($\psi=$) /u/, whereas vowel ($y=$) /u/ itself was the second lowest accurately classified group (i.e $y \neq \psi$).  \\

One hypothesis is that our EEG signals contain the components with sources other than language-related ones. This may be because of the inevitable components that originate from non-language-related areas, such as vowel imagery (when the subjects think about the shape of vowels) or emotions (when subjects experience a slight feeling with specific vowels) \cite{Idrees2016eeg, min2016vowel}. For example, vowel /a/ is a reminder of joy or anger moments but not a grief moment. A grief feeling is often be accompanied with /e/. More studies are required to test this hypothesis experimentally. 

\begin{table}[]
    \centering
    \begin{tabular}{|c||ccccc|}
    \hline 
    		&		${\mathbf {\psi}}$= /a/	&	${\mathbf {\psi}}$= /e/		&		${\mathbf {\psi}}$= /i/	&	${\mathbf {\psi}}$= /o/		&	${\mathbf {\psi}}$=/u/		\\
    		\hline	\hline	
    	  $y$= /a/	&	0.9		&	0.05	&	0.05	&	0.00	&	0.00	\\
          $y$= /e/	&	0.125	&	0.8		&	0.05	&	0.025	&	0.00	\\   
       $y$= /i/	&	0.00	&	0.15	&	0.655	&	0.12	&	0.075 \\
          $y$= /o/	&	0.00	&	0.05	&	0.15	&	0.8		&	0.00	\\
          $y$= /u/	&	0.05	&	0.15	&	0.00	&	0.125	&	0.675	\\
     \hline	
    \end{tabular}
    \caption{Confusion matrix: $y$ is the true label of the test samples $\{\bx^* \in S^* \}$, and $\psi$ is the predicted label by the proposed BCI framework on samples $\{\bx^* \in S^* \}$.}
    \label{tab:conf_mat}
\end{table}

\begin{table}[]
    \centering
    \begin{tabular}{|c|c|c|c|c|c|}
    \hline
	            & /a/	    &	/e/	    &	/i/	    &	/o/		&	/u/		\\
	\hline
    $y= \psi$    &   90\%    &   80\%    &   65.5\%    &   80\%    &   67.5\%   \\
    \hline
    $\hat{\eps}_{(\bx^*,y)}$   &  0.1     &   0.2     &   0.345    &   0.2     &   0.325    \\ 
    \hline
    \end{tabular}
    \caption{Prediction percent accuracy and the error contributed by each class ($\hat{\eps}|y$) on the test-like data set $S^*$ per class.}
    \label{tab:acc}
\end{table}

\begin{table}[]
    \centering
    \begin{tabular}{|c|cccc|c|}
    \hline 
		
    	$y=$ /a/	&	{\bf 5 (/e/)} 	&	{\bf 5 (/i/)}	&	0 ( /o/)	&	0 ( /u/) 	\\
        $y=$  /e/	&   {\bf 12.5  (/a/)} &	5 ( /i/)	&	2.5 (/o/)	&	0 (/u/)	    \\   
       $y=$  /i/	    &	0 ( /a/)	&	{\bf 15 (/e/)}	&	12 ( /o/)	&	7.5 (/u/)  \\
       $y=$  /o/	    &	0 ( /a/)	&	5 (/e/)	&	{\bf 15 (/i/)}	&	0 (/u/)	\\
       $y=$  /u/ 	&	5 ( /a/)	&	{\bf 15 (/e/)}	&	0 (/i/)	&	12.5 (/o/)	\\
     \hline	
    \end{tabular}
    \caption{Percentage of confusion with other vowels. Inside parenthesis is the computed $\psi$, i.e. the The predicted vowel by the proposed BCI.}
    \label{tab:misclass}
\end{table}

\begin{table}[]
    \centering
    \begin{tabular}{|l|l|c|}
    \hline
         Distance metric    &   description &  $\hat{\eps}_{S^*}$ \\
         \hline
            Euclidean       &   the metric used in this paper &     0.23    \\
            Seuclidean      &   Standardized Euclidean          &  0.3 \\
            Cosine          &   $ 1 - \cos (\phi)$ &   0.32 \\
            Spearman        &   $1 - \rho $ &   0.32  \\
            \hline
    \end{tabular}
    \caption{The error rate for alternative configuration of the decision tree, which were constructed by a variety of distance metrics $\hat{\eps}_{S^*}$ is the error rate on the test-like data set $S^*$, $\phi$ is the angle between averaged feature vectors of classes, $\rho$ is the Spearman's rank correlation between classes.}
    \label{tab:my_label}
\end{table}

\section{Conclusion}
We have studied the suitability of the state of the art machine learning algorithms for recognition of English vowels in EEG data as potentially part of a future BCI system. In this work several steps was taken to deal with the complexity of the EEG signals while keeping the system as simple as possible, due to the small sample size. To generalize the binary-class classification rule SVM, a dendrogram was constructed based on the Euclidean distance between group means and then used as the decision tree for a multi-class DT-SVM. DT-SVM was superior to the commonly used methods one-vs-one and one-vs-all in terms of number of required classifiers. We experimented with different distance metrics of dendrogram construction and included the one with the lowest classification error rate in this paper. We believe that this work has potentials in significantly helping people with vocal impairments, as well as automatic and easy communication of emotions between patients and healthcare providers. The experimental results showed that the vowels contribute to the accuracy of the system differently, which brought up the hypothesis that the subjects may have had incorporated unconscious tasks other than pure vowel imagination, such as imagining the shape of the vowels or experiencing a slight emotion. Further researches are needed to properly leverage the components that are inevitably included in such EEG signals, and report the ones that need to be taken into consideration. In addition, another research direction can be to evaluate the generalization of the proposed framework on a larger data set, possibly from other EEG-based BCI applications, which may pave the road for transfer learning from the well-developed domains to the poor-data domains.    

\section{Acknowledgment}
Authors would like to thank Dr. Luis Carlos Sarmiento and his group as well as Dr. Andres Tovar for providing the EEG data of this research. 

\bibliographystyle{elsarticle-num}
\bibliography{EEG-BCI.bib}

\begin{thebibliography}{10}
\expandafter\ifx\csname url\endcsname\relax
  \def\url#1{\texttt{#1}}\fi
\expandafter\ifx\csname urlprefix\endcsname\relax\def\urlprefix{URL }\fi
\expandafter\ifx\csname href\endcsname\relax
  \def\href#1#2{#2} \def\path#1{#1}\fi

\bibitem{haas2003hans}
L.~F. Haas, Hans berger (1873--1941), richard caton (1842--1926), and
  electroencephalography, Journal of Neurology, Neurosurgery \& Psychiatry
  74~(1) (2003) 9--9.

\bibitem{wan2019review}
X.~Wan, K.~Zhang, S.~Ramkumar, J.~Deny, G.~Emayavaramban, M.~S. Ramkumar, A.~F.
  Hussein, A review on electroencephalogram based brain computer interface for
  elderly disabled, IEEE Access 7 (2019) 36380--36387.

\bibitem{abiri2019comprehensive}
R.~Abiri, S.~Borhani, E.~W. Sellers, Y.~Jiang, X.~Zhao, A comprehensive review
  of eeg-based brain--computer interface paradigms, Journal of neural
  engineering 16~(1) (2019) 011001.

\bibitem{gu2020eeg}
X.~Gu, Z.~Cao, A.~Jolfaei, P.~Xu, D.~Wu, T.-P. Jung, C.-T. Lin, Eeg-based
  brain-computer interfaces (bcis): A survey of recent studies on signal
  sensing technologies and computational intelligence approaches and their
  applications, arXiv preprint arXiv:2001.11337 (2020).

\bibitem{lotte2018review}
F.~Lotte, L.~Bougrain, A.~Cichocki, M.~Clerc, M.~Congedo, A.~Rakotomamonjy,
  F.~Yger, A review of classification algorithms for eeg-based brain--computer
  interfaces: a 10 year update, Journal of neural engineering 15~(3) (2018)
  031005.

\bibitem{huang2019eeg}
H.~Huang, Q.~Xie, J.~Pan, Y.~He, Z.~Wen, R.~Yu, Y.~Li, An eeg-based brain
  computer interface for emotion recognition and its application in patients
  with disorder of consciousness, IEEE Transactions on Affective Computing
  (2019).

\bibitem{pan2016eeg}
J.~Pan, Y.~Li, J.~Wang, An eeg-based brain-computer interface for emotion
  recognition, in: 2016 international joint conference on neural networks
  (IJCNN), IEEE, 2016, pp. 2063--2067.

\bibitem{ikeda2012effect}
K.~Ikeda, T.~Higashi, K.~Sugawara, K.~Tomori, H.~Kinoshita, T.~Kasai, The
  effect of visual and auditory enhancements on excitability of the primary
  motor cortex during motor imagery: a pilot study, International Journal of
  Rehabilitation Research 35~(1) (2012) 82--84.

\bibitem{lee2019towards}
S.-H. Lee, M.~Lee, J.-H. Jeong, S.-W. Lee, Towards an eeg-based intuitive bci
  communication system using imagined speech and visual imagery, in: 2019 IEEE
  International Conference on Systems, Man and Cybernetics (SMC), IEEE, 2019,
  pp. 4409--4414.

\bibitem{krishna2020understanding}
G.~Krishna, C.~Tran, M.~Carnahan, A.~Tewfik, Understanding effect of speech
  perception in eeg based speech recognition systems, arXiv preprint
  arXiv:2006.01261 (2020).

\bibitem{zhang2020eeg}
X.~Zhang, H.~Li, F.~Chen, Eeg-based classification of imaginary mandarin tones,
  in: 2020 42nd Annual International Conference of the IEEE Engineering in
  Medicine \& Biology Society (EMBC), IEEE, 2020, pp. 3889--3892.

\bibitem{waaramaa2008monopitched}
T.~Waaramaa, A.-M. Laukkanen, P.~Alku, E.~V{\"a}yrynen, Monopitched expression
  of emotions in different vowels, Folia Phoniatrica et Logopaedica 60~(5)
  (2008) 249--255.

\bibitem{yoshimura2011usability}
N.~Yoshimura, A.~Satsuma, C.~S. DaSalla, T.~Hanakawa, M.-a. Sato, Y.~Koike,
  Usability of eeg cortical currents in classification of vowel speech imagery,
  in: 2011 International Conference on Virtual Rehabilitation, IEEE, 2011, pp.
  1--2.

\bibitem{dasalla2009single}
C.~S. DaSalla, H.~Kambara, M.~Sato, Y.~Koike, Single-trial classification of
  vowel speech imagery using common spatial patterns, Neural networks 22~(9)
  (2009) 1334--1339.

\bibitem{wang2013extending}
L.~Wang, X.~Zhang, Y.~Zhang, Extending motor imagery by speech imagery for
  brain-computer interface, in: 2013 35th Annual International Conference of
  the IEEE Engineering in Medicine and Biology Society (EMBC), IEEE, 2013, pp.
  7056--7059.

\bibitem{trunk1979problem}
G.~V. Trunk, A problem of dimensionality: A simple example, IEEE Transactions
  on pattern analysis and machine intelligence~(3) (1979) 306--307.

\bibitem{hughes1968mean}
G.~Hughes, On the mean accuracy of statistical pattern recognizers, IEEE
  transactions on information theory 14~(1) (1968) 55--63.

\bibitem{malmivuo1995bioelectromagnetism}
J.~Malmivuo, R.~Plonsey, et~al., Bioelectromagnetism: principles and
  applications of bioelectric and biomagnetic fields, Oxford University Press,
  USA, 1995.

\bibitem{Sarmiento2019recognition}
L.~C. {Sarmiento}, J.~B. {Rodríguez}, O.~{López}, S.~I. {Villamizar}, R.~D.
  {Guevara}, C.~J. {Cortes-Rodriguez}, Recognition of silent speech syllables
  for brain-computer interfaces, in: 2019 IEEE International Conference on
  E-health Networking, Application Services (HealthCom), 2019, pp. 1--5.

\bibitem{sarmiento2014brain}
L.~Sarmiento, P.~Lorenzana, C.~Cortes, W.~Arcos, J.~Bacca, A.~Tovar, Brain
  computer interface (bci) with eeg signals for automatic vowel recognition
  based on articulation mode, in: 5th ISSNIP-IEEE Biosignals and Biorobotics
  Conference (2014): Biosignals and Robotics for Better and Safer Living (BRC),
  IEEE, 2014, pp. 1--4.

\bibitem{pei2011spatiotemporal}
X.~Pei, E.~C. Leuthardt, C.~M. Gaona, P.~Brunner, J.~R. Wolpaw, G.~Schalk,
  Spatiotemporal dynamics of electrocorticographic high gamma activity during
  overt and covert word repetition, Neuroimage 54~(4) (2011) 2960--2972.

\bibitem{ghane2015silent}
P.~Ghane, Silent speech recognition in eeg-based brain computer interface,
  Ph.D. thesis (2015).

\bibitem{stoica2005spectral}
P.~Stoica, R.~L. Moses, et~al., Spectral analysis of signals (2005).

\bibitem{pearson1901liii}
K.~Pearson, Liii. on lines and planes of closest fit to systems of points in
  space, The London, Edinburgh, and Dublin philosophical magazine and journal
  of science 2~(11) (1901) 559--572.

\bibitem{shaghaghian2021studying}
Z.~Shaghaghian, F.~Shahsavari, E.~Delzendeh, Studying buildings outlines to
  assess and predict energy performance in buildings: A probabilistic approach,
  arXiv preprint arXiv:2111.08844 (2021).

\bibitem{ghane2015robust}
P.~Ghane, G.~Hossain, A.~Tovar, Robust understanding of eeg patterns in silent
  speech, in: 2015 National Aerospace and Electronics Conference (NAECON),
  IEEE, 2015, pp. 282--289.

\bibitem{shlens2014tutorial}
J.~Shlens, A tutorial on principal component analysis, arXiv preprint
  arXiv:1404.1100 (2014).

\bibitem{lotte2007review}
F.~Lotte, M.~Congedo, A.~L{\'e}cuyer, F.~Lamarche, B.~Arnaldi, A review of
  classification algorithms for eeg-based brain--computer interfaces, Journal
  of neural engineering 4~(2) (2007) R1.

\bibitem{takahashi2002decision}
F.~Takahashi, S.~Abe, Decision-tree-based multiclass support vector machines,
  in: Proceedings of the 9th International Conference on Neural Information
  Processing, 2002. ICONIP'02., Vol.~3, IEEE, 2002, pp. 1418--1422.

\bibitem{vapnik2013nature}
V.~Vapnik, The nature of statistical learning theory, Springer science \&
  business media, 2013.

\bibitem{kijsirikul2002multiclass}
B.~Kijsirikul, N.~Ussivakul, Multiclass support vector machines using adaptive
  directed acyclic graph, in: Proceedings of the 2002 International Joint
  Conference on Neural Networks. IJCNN'02 (Cat. No. 02CH37290), Vol.~1, IEEE,
  2002, pp. 980--985.

\bibitem{govindarajan2010evaluation}
M.~Govindarajan, R.~Chandrasekaran, Evaluation of k-nearest neighbor classifier
  performance for direct marketing, Expert Systems with Applications 37~(1)
  (2010) 253--258.

\bibitem{andersson1999measure}
A.~Andersson, P.~Davidsson, J.~Lind{\'e}n, Measure-based classifier performance
  evaluation, Pattern Recognition Letters 20~(11-13) (1999) 1165--1173.

\bibitem{Idrees2016eeg}
B.~M. {Idrees}, O.~{Farooq}, Eeg based vowel classification during speech
  imagery, in: 2016 3rd International Conference on Computing for Sustainable
  Global Development (INDIACom), 2016, pp. 1130--1134.

\bibitem{min2016vowel}
B.~Min, J.~Kim, H.-j. Park, B.~Lee, Vowel imagery decoding toward silent speech
  bci using extreme learning machine with electroencephalogram, BioMed research
  international 2016 (2016).

\end{thebibliography}

\end{document}